\documentclass[letterpaper, 10 pt, conference]{ieeeconf}  

\IEEEoverridecommandlockouts                              

\usepackage{cite}
\usepackage{amsmath,amssymb,amsfonts}
\usepackage{algorithmic}
\usepackage{graphicx}
\usepackage{textcomp}
\usepackage[dvipsnames]{xcolor}
\usepackage{multirow}
\usepackage{adjustbox}
\usepackage{caption}
\usepackage{subcaption}
\usepackage{balance}
\usepackage{float}
\usepackage{url}
\usepackage{hyperref}

\title{\LARGE \bf
Probabilistic 3D Multi-Object Cooperative Tracking for \\
Autonomous Driving via Differentiable Multi-Sensor Kalman Filter
}

\author{Hsu-kuang Chiu$^1$, Chien-Yi Wang$^2$, Min-Hung Chen$^2$, and Stephen F. Smith$^1$
\thanks{$^1$ Robotics Institute of Carnegie Mellon University. $^2$ NVIDIA Research.}%
\thanks{This research was funded in part by the CMU Safety21 University Transportation Center, the NSF ACCESS program, and the Google cloud research credit program.}%
}

\begin{document}

\maketitle
\thispagestyle{empty}
\pagestyle{empty}

\begin{abstract}
Current state-of-the-art autonomous driving vehicles mainly rely on each individual sensor system to perform perception tasks. Such a framework's reliability could be limited by occlusion or sensor failure. To address this issue, more recent research proposes using vehicle-to-vehicle (V2V) communication to share perception information with others. However, most relevant works focus only on cooperative detection and leave cooperative tracking an underexplored research field. A few recent datasets, such as V2V4Real, provide 3D multi-object cooperative tracking benchmarks. However, their proposed methods mainly use cooperative detection results as input to a standard single-sensor Kalman Filter-based tracking algorithm. In their approach, the measurement uncertainty of different sensors from different connected autonomous vehicles (CAVs) may not be properly estimated to utilize the theoretical optimality property of Kalman Filter-based tracking algorithms. In this paper, we propose a novel 3D multi-object cooperative tracking algorithm for autonomous driving via a differentiable multi-sensor Kalman Filter. Our algorithm learns to estimate measurement uncertainty for each detection that can better utilize the theoretical property of Kalman Filter-based tracking methods. The experiment results show that our algorithm improves the tracking accuracy by 17\% with only 0.037x communication costs compared with the state-of-the-art method in V2V4Real. Our code and videos are available at the \href{https://github.com/eddyhkchiu/DMSTrack}{URL} and the \href{https://eddyhkchiu.github.io/dmstrack.github.io/}{URL}.

\end{abstract}

\section{Introduction}
Autonomous driving vehicles have been deployed for commercial usage in certain regions of the world due to the significant advance of the relevant technologies over the past decade. However, currently deployed large-scale autonomous driving systems mainly rely on the individual autonomous vehicle's perception system. Such a system could have difficulty detecting and tracking other vehicles, pedestrians, cyclists, or any type of vulnerable road users due to occlusion or sensor failure. To address such safety concerns, more and more research datasets~\cite{xu2023v2v4real, li2022v2xsim, yu2023v2xseq} and algorithms~\cite{chen2019fcooper, xu2022opencood, wang2020v2vnet, xu2022v2xvit, xu2022cobevt, li2021disco, liu2020when2com, liu2020who2com} for cooperative perception have been published in recent years. In cooperative perception for autonomous driving, each connected autonomous vehicle (\textbf{CAV}) can share its own raw sensor input (\textbf{early fusion}), encoded perception features (\textbf{intermediate fusion}), or detection results (\textbf{late fusion}) with others via vehicle-to-vehicle (V2V) communication in order to achieve better overall perception coverage and accuracy.
For the most part, work in cooperative perception has focused on cooperative detection \cite{chen2019fcooper, xu2022opencood, wang2020v2vnet, xu2022v2xvit, xu2022cobevt, li2021disco, liu2020when2com, liu2020who2com}, proposing different types of \textbf{intermediate fusion} cooperative detection algorithms and demonstrating promising improvements in detection accuracy over both \textbf{no fusion} and \textbf{late fusion} baseline methods.

\begin{figure}[]
    \centering
    \begin{subfigure}[b]{0.23\textwidth}
        \centering 
        \includegraphics[width=\textwidth]{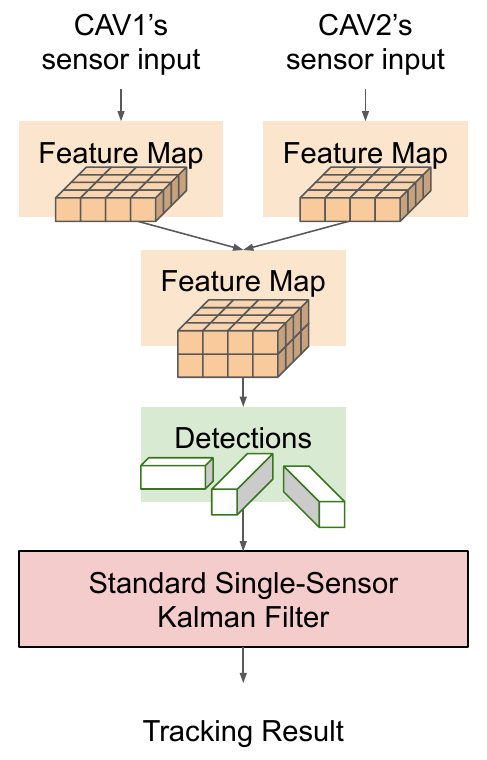}
        \caption[]%
        {{V2V4Real~\cite{xu2023v2v4real}}}    
        \label{fig:baseline}
    \end{subfigure}
    \hfill
    \begin{subfigure}[b]{0.23\textwidth}  
        \centering 
        \includegraphics[width=\textwidth]{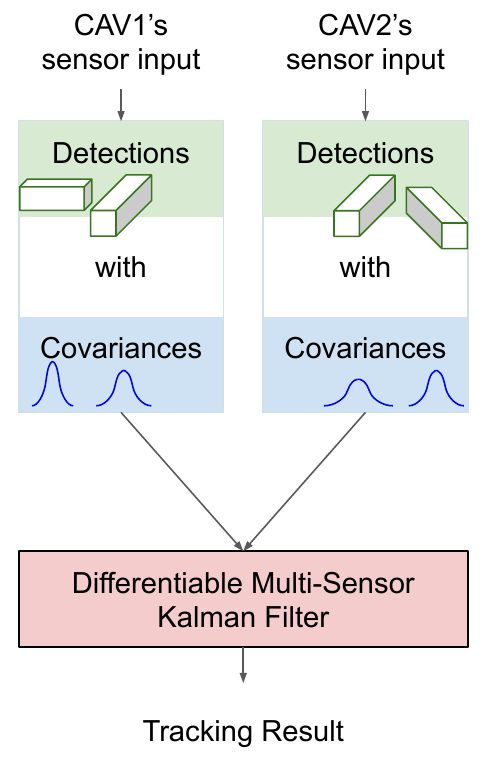}
        \caption[]%
        {{DMSTrack (ours)}}    
        \label{fig:ours}
    \end{subfigure}
    \caption[]
    {Cooperative tracking architecture comparison. 
    \vspace{-15pt}} 
    \label{fig:comparison}
\end{figure}

Much less attention to date has been given to the problem of 3D multi-object cooperative tracking. 
Some datasets, such as V2V4Real~\cite{xu2023v2v4real}, do provide 3D multi-object cooperative tracking benchmarks. But these benchmarks typically apply the aforementioned cooperative detection algorithms and then use the detection results as input to typical 3D multi-object tracking algorithms derived from the standard single-sensor Kalman Filter~\cite{kalman1960filter}, as shown in Figure \ref{fig:baseline}. The measurement uncertainty of different sensors from different CAVs may not be properly estimated to utilize the theoretical optimality property of Kalman Filter-based tracking algorithms.

To address this problem, we propose a novel algorithm: Differentiable Multi-Sensor Kalman Filter for 3D Multi-Object Cooperative Tracking (\textbf{DMSTrack}), as shown in Figure \ref{fig:ours} and Figure \ref{fig:architecture}. Our algorithm is designed to be capable of \textbf{estimating observation noise covariance of each detection from different CAVs to better take advantage of the Kalman Filter's theoretical optimality property}: minimizing the expected error of state estimation. 

To achieve such an algorithm design goal, we reconsider the aforementioned cooperative perception fusion approaches, and determine that, due to its extensibility, the \textbf{late fusion} approach particularly fits our algorithm design principle. In this approach, each CAV first performs its individual object detection using its own sensor input. So we can easily reformulate the cooperative tracking problem in this approach as a state estimation problem in a multi-sensor system and design our algorithm upon the foundation of prior works in this field. The original multi-sensor Kalman Filter ~\cite{willner1976multi_sensor_kalman_filter} work focused on the single-object tracking problem in a multi-sensor system. This research proves that sequentially processing each sensor's measurement on the same object with multiple Kalman Filter's update steps in a discrete timestep also achieves theoretically optimal state estimation results as long as the following precondition holds: each sensor's measurement is synchronized and statistically independent. Inspired by this idea, our first innovation is to extend \cite{willner1976multi_sensor_kalman_filter} to have multi-object tracking capability by applying multiple data associations and Kalman Filter update steps to build our cooperative tracking algorithm, as shown in Figure \ref{fig:architecture}. Although the theoretical guarantee property of \cite{willner1976multi_sensor_kalman_filter} is for single-object tracking and the precondition may not always hold in cooperative perception driving datasets, our experiment still shows that our proposed new tracking algorithm provides promising performance improvement in comparison to the baseline methods in V2V4Real~\cite{xu2023v2v4real}.

The second innovation of our proposed cooperative tracking algorithm is its ability to estimate the observation noise covariance of each detection from different CAVs. The previously developed Backprop Kalman Filter~\cite{haarnoja2016backpropkf} proposes the use of a deep neural network to learn a single observation noise covariance matrix from images for the 2D single-object tracking problem in videos and the visual odometry problem. Compared to these problems, 3D multi-object cooperative tracking for autonomous driving is a much more complex problem. The measurement uncertainty of different bounding box variables of each detection could be related to the sensor system, the 3D object detection model, and the relative pose of the detected object to the sensor. Thus, we propose using local point cloud features and positional features as input to our designed covariance neural network to learn the observation noise covariance. We use the regression loss between the tracking result boxes and the ground-truth boxes to train our proposed differentiable multi-sensor Kalman Filter, as shown in Figure \ref{fig:architecture} and Figure \ref{fig:model}. This approach further improves the cooperative tracking performance in the V2V4Real\cite{xu2023v2v4real} dataset. 

In terms of communication cost, our approach only needs to share bounding box and covariance information for each detection as shown in Figure \ref{fig:ours}. So it has a much smaller communication cost compared with the \textbf{intermediate fusion} baseline methods in V2V4Real~\cite{xu2023v2v4real}, which need to share scene-level Bird-Eye-View (BEV) feature maps with other CAVs as shown in Figure \ref{fig:baseline}.

In summary, we propose a novel probabilistic 3D multi-object cooperative tracking algorithm for CAVs via differentiable multi-sensor Kalman Filter: \textbf{DMSTrack}. Our model learns to estimate the covariance matrix for each detection from each individual CAV. The detection results with covariances are merged via multiple data associations and Kalman Filter update steps to generate the final cooperative tracking results in each timestep. We evaluate our proposed algorithm's performance in the V2V4Real~\cite{xu2023v2v4real} dataset, which is the first large-scale worldwide available cooperative perception real dataset for autonomous driving with 3D multi-object cooperative tracking benchmarks. Our proposed algorithm, \textbf{DMSTrack}, improves the tracking accuracy by 17\% with only 0.037x communication costs compared with the state-of-the-art method CoBEVT~\cite{xu2022cobevt} in V2V4Real~\cite{xu2023v2v4real}.

\section{Related Work}

\subsection{Single-Agent Autonomous Driving}
Single-agent autonomous driving large-scale datasets~\cite{geiger2012kitti, caesar2019nuscenes,  sun2020waymo, ettinger2021waymo} and algorithms~\cite{lang2019pointpillar, yin2020center, weng2019ab3dmot, weng2020gnn3dmot, chiu2020probabilistic, chiu2021probabilistic, hu2023uniad} serve as  important foundations for cooperative autonomous driving research. For example, we follow V2V4Real~\cite{xu2023v2v4real}'s late fusion approach to use PointPillar~\cite{lang2019pointpillar} as the single-agent 3D object detector in our experiments. Our algorithm also adopts some components from AB3DMOT~\cite{weng2019ab3dmot}, such as the data association criteria and the track life cycle management.

\subsection{Cooperative Perception with CAVs}
Cooperative perception for autonomous driving has gained more attention in the past few years. F-Cooper~\cite{chen2019fcooper} is the first work that introduces feature-level fusion to improve CAVs' cooperative detection accuracy. V2VNet~\cite{wang2020v2vnet} adopts graph neural networks to aggregate intermediate features. AttFuse~\cite{xu2022opencood}, V2X-ViT~\cite{xu2022v2xvit}, and CoBEVT~\cite{xu2022cobevt} develop varieties of attention and transformer-based models for cooperative detection.
V2V4Real~\cite{xu2023v2v4real} is the first worldwide available public real-world dataset with 3D cooperative detection and tracking benchmarks, and it mainly applies the aforementioned cooperative detection algorithms' results as input to a standard single-sensor Kalman Filter-based tracking algorithm: AB3DMOT~\cite{weng2019ab3dmot}. In addition to V2V4Real~\cite{xu2023v2v4real}, a simulation dataset V2X-Sim~\cite{li2022v2xsim} and a real dataset V2X-Seq~\cite{yu2023v2xseq} also provide cooperative tracking benchmarks. However, V2X-Sim~\cite{li2022v2xsim}'s tracking benchmark degrades detection and tracking evaluation to BEV xy-axis-aligned 2D bounding box format without properly considering the center positions in the z-axis, heights, and orientations. V2X-Seq~\cite{yu2023v2xseq} is a new cooperative perception dataset but is not available worldwide. Therefore, evaluation on these two datasets is out of the scope of our current research. We focus on evaluating our algorithm in the V2V4Real~\cite{xu2023v2v4real} benchmark. Different from the benchmark's best-performing method CoBEVT~\cite{xu2022cobevt} which is an intermediate fusion approach, our proposed method is a late fusion approach with better tracking accuracy and a lower communication cost.

\subsection{Kalman Filters}
The multi-sensor Kalman Filter~\cite{willner1976multi_sensor_kalman_filter} provides the theoretical foundation of our algorithm. Backprop Kalman Filter~\cite{haarnoja2016backpropkf} inspires our neural network approach of learning covariance from sensor input to better take the advantage of the theoretical optimality property from~\cite{willner1976multi_sensor_kalman_filter}.
Recent works~\cite{lee2020multimodal, kloss2021how} applying Backprop Kalman Filter~\cite{haarnoja2016backpropkf} only focus on single object tracking. To the best of our knowledge, we are the first to integrate both \cite{willner1976multi_sensor_kalman_filter} and \cite{haarnoja2016backpropkf} to create the differentiable multi-sensor Kalman Filter model and extend it to build our probabilistic 3D multi-object cooperative tracking algorithm for autonomous driving.

\section{Method}
\begin{figure}[]
\centering
\includegraphics[width=0.48\textwidth]{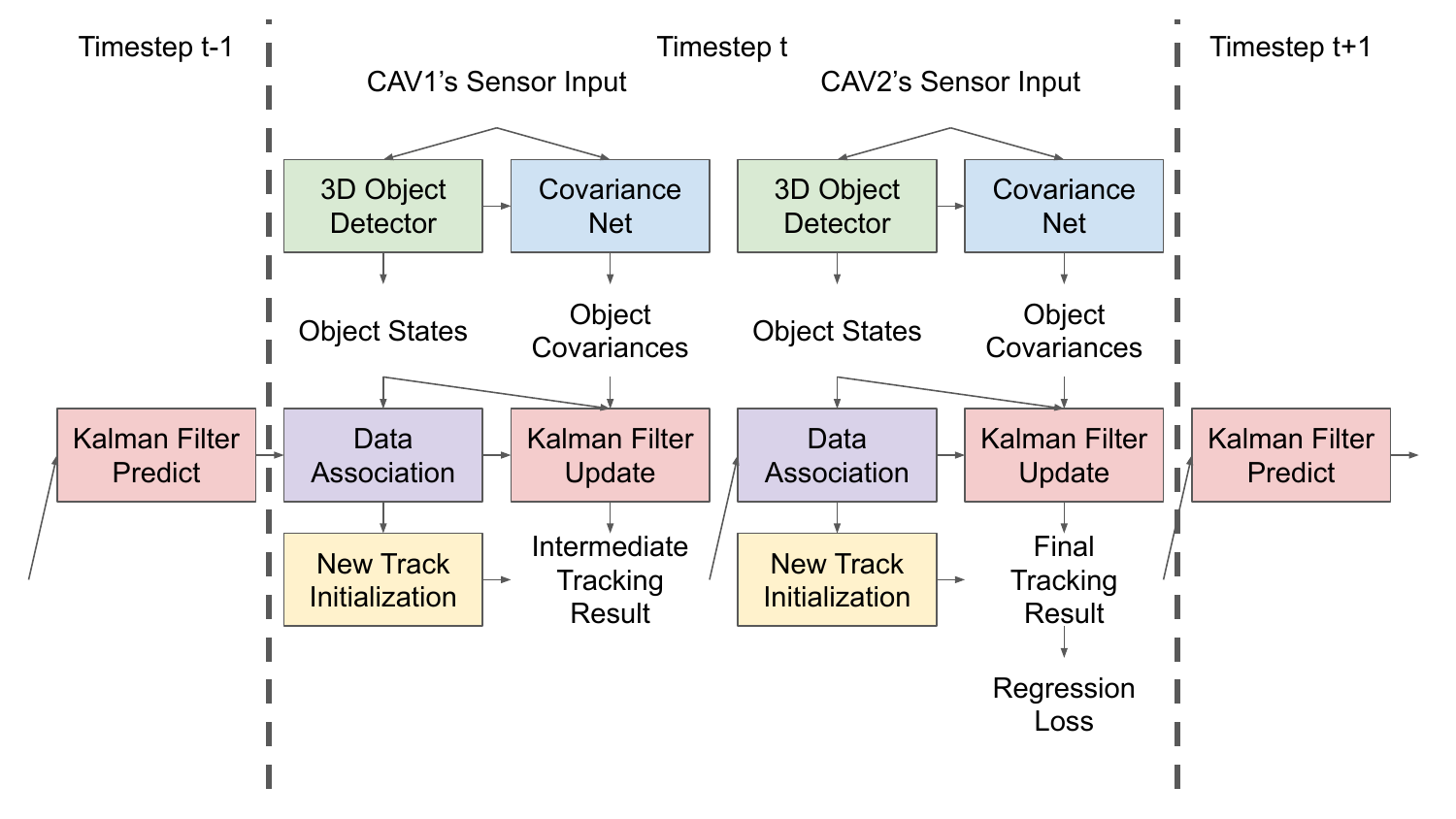}
\caption[]
        {Architecture diagram of our proposed cooperative tracking algorithm applied to a minimal example with two CAVs.} 
        \vspace{-15pt}
        \label{fig:architecture}
\end{figure}

Our proposed cooperative tracking architecture is illustrated in Figure \ref{fig:architecture} with a minimal example of two CAVs. At each timestep $t$, each CAV first feeds its own sensor input to a 3D object detector and our designed covariance neural network to detect object state means and estimate object state covariances in a decentralized manner. Then each CAV transforms its detected object states from its own local coordinate system to a global coordinate system and transmits the detection information to a centralized agent that maintains our proposed differentiable multi-sensor Kalman Filters to track multiple object states and covariances in the global coordinate system. The centralized agent can be any CAV or an external computing node.


The centralized agent then sequentially performs multiple data associations and Kalman Filter update steps with each CAV's detected object states and covariances, as shown in Figure \ref{fig:architecture}. Once the centralized agent finishes processing the detection information from all CAVs, the latest tracking result is treated as the final cooperative tracking results at this timestep $t$. A Kalman Filter prediction step is then performed to predict the object states for the next timestep given the Kalman Filter's process model.


In the following subsections, we describe more details of our proposed differentiable multi-sensor Kalman Filter setting for 3D multi-object cooperative tracking, followed by the covariance neural network's input feature formation, model architecture, loss function, and training strategy.

\subsection{Kalman Filter Setting}
\subsubsection{Object State}
To apply Kalman Filters \cite{kalman1960filter} to the 3D multi-object cooperative tracking for autonomous driving problem, we follow V2V4Real\cite{xu2023v2v4real} and AB3DMOT\cite{weng2019ab3dmot} to represent a track state at timestep $t$ by a vector of $10$ variables as follows:
\begin{equation}\label{eq:state}
{\bf s}_t = (x_t, y_t, z_t, a_t, l_t, w_t, h_t, dx_t, dy_t, dz_t)^T,
\end{equation}
where $(x_t, y_t, z_t)$ is the center position of the bounding box, $a_t$ is the rotation angle along the z-axis, $(l_t, w_t, h_t)$ represents the length, width, and height of the bounding box. And $(dx_t, dy_t, dz_t)$ contains the velocity information, which is used to predict the track state in the next timestep. 

\subsubsection{Prediction Step}
In the Kalman Filter's prediction step, we use a constant velocity model as the process model. 


The Kalman Filter's prediction step can be represented in the following matrix form:
\begin{align}
\label{eq:predicted_mean}
    {\bf \hat{\mu}}_{t} &= {\bf A} \mathbf{\mu}_{t-1}\\
    \label{eq:predicted_covariance}
    \hat{\Sigma}_{t} & = {\bf A} \Sigma_{t-1} {\bf A}^T + {\bf Q}
\end{align}
where $\hat{\mu}_{t} \in \mathbb{R}^{10}$ is the predicted track state mean at timestep $t$, and ${\bf \mu}_{t-1} \in \mathbb{R}^{10}$ is the estimated mean of the track state ${\bf s}_{t-1}$ at timestep $t-1$. The matrix ${\bf A} \in \mathbb{R}^{10 \times 10}$ is the state transition matrix of the Kalman Filter's process model. The matrix $\hat{\Sigma}_{t} \in \mathbb{R}^{10 \times 10}$ is the predicted state covariance at timestep $t$, and $\Sigma_{t-1} \in \mathbb{R}^{10 \times 10}$ is the estimated state covariance at timestep $t-1$. The matrix ${\bf Q} \in \mathbb{R}^{10 \times 10}$ is the process model noise covariance. We follow V2V4Real\cite{xu2023v2v4real} and AB3DMOT\cite{weng2019ab3dmot} and use the same constant values for ${\bf Q}$.

\subsubsection{Data Association and Update Step}
At timestep $t$, each track could have zero, one, or more associate detections from multiple CAVs. To have a fair comparison, our algorithm's detection part uses V2V4Real~\cite{xu2023v2v4real}'s \textbf{late fusion} baseline's PointPillar~\cite{lang2019pointpillar} model checkpoint as the single-agent 3D object detector. Each detection is represented by $7$ variables used as observation in the Kalman Filter as follows:
\begin{align}
{\bf o}_t = (x_t, y_t, z_t, a_t, l_t, w_t, h_t)^T, \label{eq:observation}
\end{align}

Our proposed algorithm then sequentially processes all CAV's detections with covariances via multi-sensor Kalman Filters with multiple data associations and update steps, as shown in Figure \ref{fig:architecture}. For data association, we follow V2V4Real\cite{xu2023v2v4real} and AB3DMOT\cite{weng2019ab3dmot} to use 3D Intersection-over-Union (3D IOU) as the similarity metric and Hungarian algorithm~\cite{kuhn1955hungarian}. For each matched pair of a track and a detection, the Kalman Filter's update step works as follows:
\begin{align}
    {\bf S}_{t} & = {\bf H} \hat{\Sigma}_{t} {\bf H}^T + {\bf R} \label{eq:S} \\ 
    {\bf K}_{t} & = \hat{\Sigma}_{t} {\bf H}^T {\bf S}_{t}^{-1}\\
    \label{eq:estimated_mean} \mu_{t} & = \hat{\mu}_{t} + {\bf K}_{t} ({\bf o}_{t} - {\bf H} \hat{\mu}_{t})  \\
    \label{eq:estimated_covariance} \Sigma_{t} & = ({\bf I} - {\bf K}_{t}{\bf H}) \hat{\Sigma}_{t},  
\end{align}
where ${\bf S}_{t} \in \mathbb{R}^{7 \times 7}$ is the innovation covariance matrix, ${\bf H} = [{\bf I}~{\bf 0} ] \in \mathbb{R}^{7 \times 10}$ is the observation model matrix, $\hat{\Sigma}_{t} \in \mathbb{R}^{10 \times 10}$ is the predicted state covariance from equation (\ref{eq:predicted_covariance}), $ {\bf R} \in \mathbb{R}^{7 \times 7}$ is the observation noise covariance matrix for the detection, ${\bf K}_{t} \in \mathbb{R}^{10 \times 7}$ is the Kalman Gain, $\mu_{t} \in \mathbb{R}^{10}$ is the estimated state mean at timestep $t$ as the tracking result, $\hat{\mu}_{t} \in \mathbb{R}^{10}$ is the predicted  state mean from equation (\ref{eq:predicted_mean}), ${\bf o}_{t} \in \mathbb{R}^{7}$ is the detection from equation (\ref{eq:observation}), $\Sigma_{t} \in \mathbb{R}^{10 \times 10}$ is the estimated state covariance at timestep $t$, and ${\bf I}$ represents an identity matrix. In our algorithm, we assume all observation noise covariance ${\bf R} \in \mathbb{R}^{7 \times 7}$ and track initial state covariance $\Sigma_0  \in \mathbb{R}^{10 \times 10}$ matrices are all diagonal.

For unmatched detections and tracks, we perform the same track life cycle management approach as in V2V4Real\cite{xu2023v2v4real} and AB3DMOT~\cite{weng2019ab3dmot}. Our algorithm repeats the update step after each data association to sequentially process the detection results of all CAVs and generates the final tracking results, as shown in Figure \ref{fig:architecture}.

\subsection{Input Feature Formation}
To learn the observation noise covariance ${\bf R} \in \mathbb{R}^{7 \times 7}$ in equation (\ref{eq:S}) for every detection, we use two types of input features: the local BEV feature and the positional feature.

\subsubsection{Local BEV Feature}
Each detection's local BEV feature is extracted from the PointPillar\cite{lang2019pointpillar} 3D object detector's BEV feature maps. Given an object detection's center position, we first find the corresponding cell in two BEV feature maps right before and after PointPillar\cite{lang2019pointpillar}'s 2D CNN backbone. We extract two local BEV features respectively and merge them to generate this detection's local BEV feature ${\bf F}_{bev} \in \mathbb{R}^{320 \times 20 \times 20}$.

\subsubsection{Positional Feature}
Each detection's observation noise covariance could also be relevant to the object's relative position and orientation to the global coordinate system and the corresponding sensor CAV's local coordinate system. The coordinate transformation matrix between the CAV's local coordinate system and the global coordinate system could also provide useful information. Therefore, we propose the following novel positional feature extraction algorithm to capture the geometry information. First, we extract a positional feature vector ${\bf f}_{pos}$ with $18$ variables for each detection as follows:
\begin{align}
  {\bf f}_{global} &= (x, y, z, a, l, w, h, \sqrt{x^2 + y^2})_{global}^T \\
  {\bf f}_{local} &= (x, y, z, a, \sqrt{x^2 + y^2})_{local}^T \\
  {\bf f}_{local\_to\_global} &= (t_x, t_y, t_z, \theta_{yaw}, \sqrt{t_x^2 + t_y^2})^T \\
  {\bf f}_{pos} &= {\bf f}_{global} \odot  {\bf f}_{local} \odot {\bf f}_{local\_to\_global},
\end{align}
where ${\bf f}_{global} \in \mathbb{R}^8$ consists of the detection variables, same as equation (\ref{eq:observation}), appended by its 2D distance to the origin of the global coordinate system. The vector ${\bf f}_{local} \in \mathbb{R}^5$ consists of the detection position, orientation, and 2D distance to the origin in the corresponding CAV's local coordinate system. The vector ${\bf f}_{local\_to\_global} \in \mathbb{R}^5$ is derived from the coordinate transformation matrix between the corresponding CAV's local coordinate system and the global coordinate system, where $(t_x, t_y, t_z)^T$ represents the translation vector and $\theta_{yaw}$ is the rotation angle along the z-axis. Finally, ${\bf f}_{pos} \in \mathbb{R}^{18}$ is the positional feature vector of this detection and the notation $\odot$ represents the concatenation operator.

Once we have the positional feature vector ${\bf f}_{pos}$, we then generate the positional feature matrix ${\bf F}_{pos}$ using our modified version of positional encoding~\cite{vaswani2017attention}. 
In our algorithm, each element $x$ in ${\bf f}_{pos}$ is transformed to a positional encoded vector ${\bf x}$ by the following equations:
\begin{align}
  \Bar{x} &= normalize(x) \\
  {\bf x}_{2i} &= sin(\frac{\Bar{x}}{2^{\frac{i}{d}}}) ~\forall ~i=0,1, ... ,d-1 \\
  {\bf x}_{2i+1} &= cos(\frac{\Bar{x}}{2^{\frac{i}{d}}})  ~\forall ~i=0,1, ... ,d-1,
\end{align}
where $\Bar{x} \in [-\pi, \pi]$ is the normalized value of $x$ by considering the min and max possible values of this variable, $d = 128$ is a constant number, ${\bf x} \in \mathbb{R}^{256}$ is the positional encoded vector, the subscripts $2i$ and $2i+1$ represent the indices of even and odd elements of the vector. We use the same approach to encode each element in ${\bf f}_{pos}$ and generate the final positional feature matrix ${\bf F}_{pos}$ as follows:
\begin{align}
  {\bf F}_{pos} &= PositionalEncoding({\bf f}_{pos}),
\end{align}
where ${\bf F}_{pos}  \in \mathbb{R}^{18 \times 256}$ is the final positional feature of this detection.

\subsection{Covariance Neural Network}

\begin{figure}[]
\centering
\includegraphics[width=0.48\textwidth]{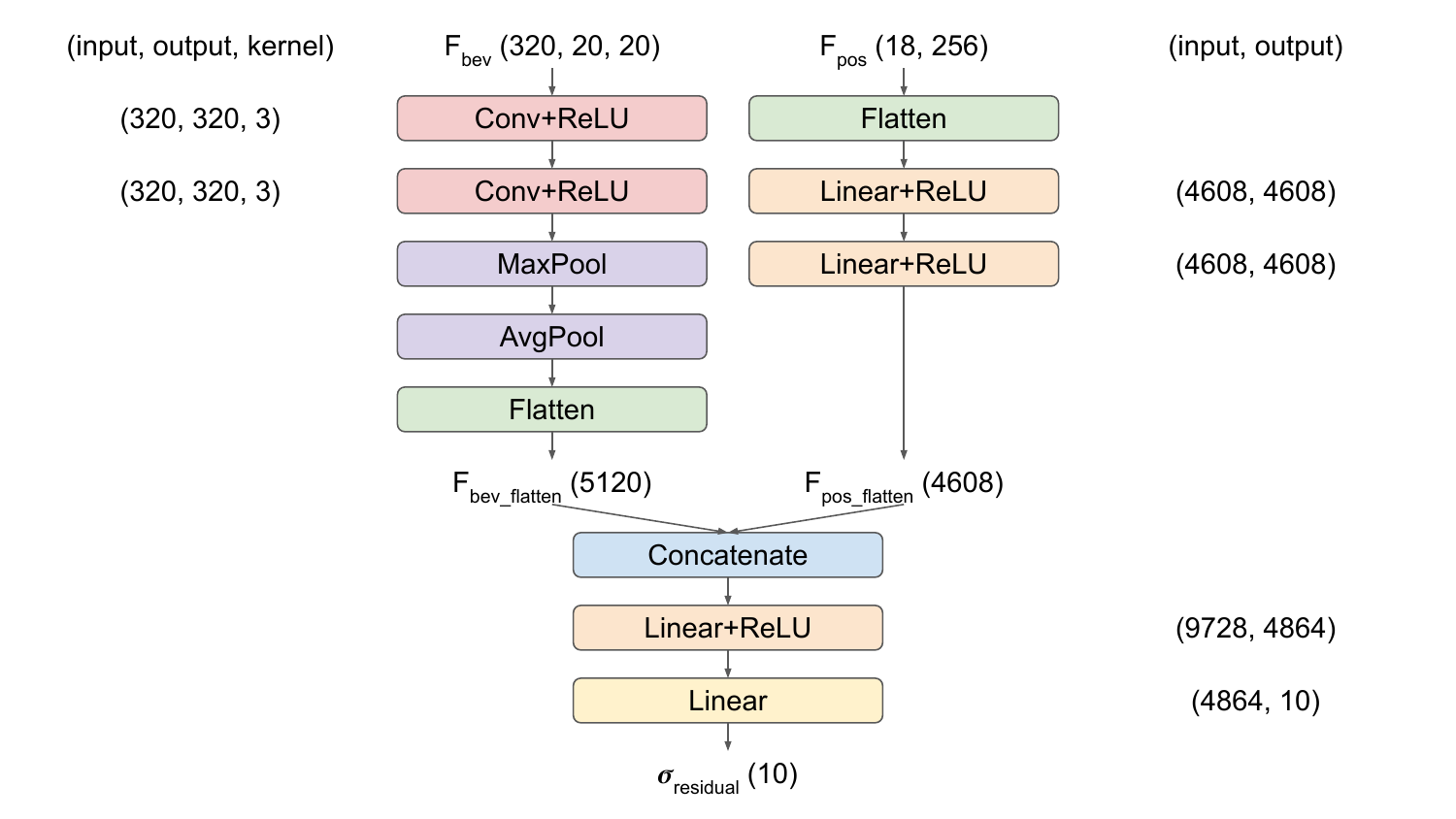}
\caption[]
        {Covariance Neural Network. The numbers represent features' tensor shapes and trainable layers' input, output, and kernel sizes.} 
        \vspace{-15pt}
        \label{fig:model}
\end{figure}

Our covariance neural network takes each detection's local BEV feature ${\bf F}_{bev}$ and positional feature ${\bf F}_{pos}$ as input and generates output to estimate this detection's covariance, as shown in Figure \ref{fig:model}. The estimated covariance is used to generate the observation noise covariance ${\bf R} \in \mathbb{R}^{7 \times 7}$ during the Kalman Filter's update step in equation (\ref{eq:S}) or the state covariance matrix $\Sigma_0 \in \mathbb{R}^{10 \times 10}$ for a newly initialized track.

The input features ${\bf F}_{bev} \in \mathbb{R}^{320 \times 20 \times 20}$ and ${\bf F}_{pos}  \in \mathbb{R}^{18 \times 256}$ have very different numbers of elements. To better balance the contributions from these two features during training, we first use the same number of trainable layers to process them individually and make the processed flatten features ${\bf F}_{bev\_flatten} \in \mathbb{R}^{5120}$ and ${\bf F}_{pos\_flatten} \in \mathbb{R}^{4608}$ more equivalent in size.

Then we concatenate these two processed flatten features and use two linear layers with a rectified linear unit to generate the final network output. Instead of directly generating the covariance values, the network generates the residuals of standard deviations in order to make the training more numerically stable as follows:
\begin{align}
    diag({\bf R}) = (sqrt(diag({\bf R}_{def})) + {\bf \sigma}_{residual}[1:7])^2, \label{eq:learnable_R}
\end{align}
where $diag$ is the operator that accesses diagonal elements of a matrix, ${\bf R} \in \mathbb{R}^{7 \times 7}$ is the detection's observation noise covariance matrix used in the Kalman Filter update step's equation (\ref{eq:S}), $sqrt$ represent the element-wise square root function, ${\bf R}_{def}$ is the constant default observation noise covariance matrix set to be an identity matrix, $ {\bf \sigma}_{residual} \in \mathbb{R}^{10}$ is the covariance neural network output, the notation $[1:7]$ means accessing the first $7$ elements of a vector, and the superscript $2$ represents the element-wise square operator.

Similarly, when a new track is initialized from an unmatched detection, we set the track's initial state covariance matrix as follows:
\begin{align}
    diag(\Sigma_0) = (sqrt(diag(\Sigma_{0_{def}})) + {\bf \sigma}_{residual})^2, \label{eq:learnable_state_covariance}
\end{align}
where $\Sigma_0 \in \mathbb{R}^{10 \times 10}$ is the state covariance matrix of a newly initialized track, $\Sigma_{0_{def}}$ is the constant default state covariance matrix set to be an identity matrix.

\subsection{Loss Function and Training Strategy}
To train the covariance neural network, we divide the training set sequences into sub-sequences, each of which has a length $T = 10$ timesteps. We run our proposed cooperative tracking algorithm on each sub-sequence and calculate the loss, as shown in Figure \ref{fig:architecture}. When calculating the loss, we only consider a subset of tracks whose closest ground-truth object is within $2$ meters. Assuming we have $N_t$ tracks satisfy this condition at timestep $t$, and we use L2-norm to calculate the loss as follows:
\begin{align}
    L = \frac{1}{\sum_{t=1}^T N_t}  \sum_{t=1}^T \sum_{i=1}^{N_t} ||(\mu_t^i[1:7] - gt_t^i)||_2 ,
\end{align}
where $L \in \mathbb{R}$ is the loss, $T = 10$ is the length of each training sub-sequence, $N_t$ is the number of tracks satisfy the condition, $\mu_t^i[1:7] \in \mathbb{R}^{7}$ represents the first $7$ variables of the $i$th track's state mean $\mu_t^i \in \mathbb{R}^{10}$ at timestep $t$, and $gt_t^i \in \mathbb{R}^{7}$ represents the track $\mu_t^i$'s closest ground-truth object variables. During training, this loss $L$ can back-propagate through every timestep $t \in [1, T]$ to all covariance neural networks used by all CAVs via our proposed differentiable multi-sensor Kalman Filter.
We use Adam\cite{kingma2015adam} optimizer to train our model, with an initial learning rate of $0.001$ and a weight decay of $0.00001$. We clip the gradient norm to $1$ and train our model for $20$ epochs on the training split. During our development, we found that clipping the gradient norm is particularly important to avoid the exploding gradient problem and make the training of our proposed differentiable multi-sensor Kalman Filter numerically stable.

\section{Experimental Results}

\subsection{Datasets}
We train our model and evaluate our cooperative tracking algorithm on the V2V4Real\cite{xu2023v2v4real} dataset, which is the first worldwide available public real-world dataset for vehicle-to-vehicle perception with 3D tracking benchmarks. 
Each of the training, validation, and testing split has a total of $14,210$, $2,000$, and $3,986$ frames of data respectively from two vehicles driven simultaneously together. The data released to the public and used in V2V4Real~\cite{xu2023v2v4real} paper's benchmarks are the LiDAR point clouds and ground-truth annotations of the training and testing splits. The training split has $32$ driving sequences and a total of $7105$ timesteps of data, and the testing split has $9$ driving sequences and a total of $1993$ timesteps of data. The frame rate is $10$Hz.


\subsection{Evaluation Metrics}
We use the same evaluation metrics from V2V4Real~\cite{xu2023v2v4real} and AB3DMOT~\cite{weng2019ab3dmot}. The primary metric is the Average Multi-Object Tracking Accuracy (AMOTA).
This AMOTA metric considers detection confidence scores and numbers of true positives, false positives, false negatives, and identity switches in a spectrum of different recall levels. 
Secondary evaluation metrics include Average Multi-Object Tracking Precision (AMOTP), scaled Average Multi-Object Tracking Accuracy (sAMOTA), Multi-Object Tracking Accuracy (MOTA), Mostly Tracked Trajectories (MT), and Mostly Lost Trajectories (ML). We use AB3DMOT~\cite{weng2019ab3dmot}'s default evaluation criteria: 3D Intersection-over-Union with a threshold value of $0.25$ when calculating the metrics.

\subsection{Quantitative Results}

\subsubsection{Tracking Performance}
Our proposed cooperative tracking algorithm (DMSTrack)'s performance can be seen in Table \ref{tab:test_results}.
Since V2V4Real~\cite{xu2023v2v4real} only releases its cooperative detection code and mentions that it uses AB3DMOT~\cite{weng2019ab3dmot} tracking algorithm without releasing its tracking code so far (as of ICRA2024 submission deadline), we implement our own V2V4Real\cite{xu2023v2v4real} baseline by integrating CoBEVT~\cite{xu2022cobevt} model checkpoint's cooperative detection results as the input to AB3DMOT~\cite{weng2019ab3dmot}'s tracking code to have a fair comparison with our algorithm. From Table \ref{tab:test_results}, we can see that our implementation of CoBEVT~\cite{xu2022cobevt} baseline has higher AMOTA than the V2V4Real~\cite{xu2023v2v4real} paper shows. More importantly, our proposed cooperative tracking method (DMSTrack) significantly outperforms our implemented CoBEVT~\cite{xu2022cobevt} baseline in AMOTA by $6.36$, which is a  $17\%$ relative improvement.

\begin{table*}[t!]
\small
\caption{Cooperative tracking performance in the testing split of V2V4Real~\cite{xu2023v2v4real} dataset in comparison with baseline methods. The baseline methods' results are from the V2V4Real~\cite{xu2023v2v4real} paper. In each column, the best-obtained results are typesetted in boldface. (* Our baseline implementation of applying CoBEVT~\cite{xu2022cobevt} detection results as input to AB3DMOT\cite{weng2019ab3dmot} tracking algorithm. Our code is available at this \href{https://github.com/eddyhkchiu/DMSTrack}{URL}.)
\vspace{-10pt}}
\label{tab:test_results}
\begin{center}
\begin{tabular}{ l ccccccc}
  \hline
  \hline
  Method & AMOTA $\uparrow$ & AMOTP $\uparrow$ & sAMOTA $\uparrow$ & MOTA $\uparrow$ & MT $\uparrow$ & ML $\downarrow$ & Cost (MB) $\downarrow$ \\
  \hline
  \hline
  No Fusion         & 16.08 & 41.60 & 53.84 & 43.46 & 29.41 & 60.18 & \textbf{0} \\
  Late Fusion       & 29.28 & 51.08 & 71.05 & 59.89 & 45.25 & 31.22 & 0.003 \\
  Early Fusion      & 26.19 & 48.15 & 67.34 & 60.87 & 40.95 & 32.13 & 0.96 \\
  \hline
  F-Cooper~\cite{chen2019fcooper}          & 23.29 & 43.11 & 65.63 & 58.34 & 35.75 & 38.91 & 0.20 \\
  AttFuse~\cite{xu2022opencood}           & 28.64 & 50.48 & 73.21 & 63.03 & 46.38 & 28.05 & 0.20 \\
  V2VNet~\cite{wang2020v2vnet}            & 30.48 & 54.28 & 75.53 & 64.85 & 48.19 & 27.83 & 0.20 \\
  V2X-ViT~\cite{xu2022v2xvit}           & 30.85 & 54.32 & 74.01 & 64.82 & 45.93 & 26.47 & 0.20 \\
  CoBEVT~\cite{xu2022cobevt}            & 32.12 & 55.61 & 77.65 & 63.75 & 47.29 & 30.32 & 0.20 \\
  \hline
  CoBEVT~\cite{xu2022cobevt} (*)        & 37.16 & 57.20 & 84.54 & 84.14 & 57.07 & 15.83 & 0.20 \\
  \hline
  DMSTrack (ours)   & \textbf{43.52} & \textbf{57.94} & \textbf{91.50} & \textbf{88.32} & \textbf{68.35} & \textbf{13.19} & 0.0073 \\
  \hline
\end{tabular}
\end{center}
\vspace{-10pt}
\end{table*}

\begin{table*}[t!]
\small
\caption{Ablation analysis on variants of our cooperative tracking algorithm.
\vspace{-10pt}}
\label{tab:test_ablation}
\begin{center}
\begin{tabular}{ l ccccccc}
  \hline
  \hline
  Method & AMOTA $\uparrow$ & AMOTP $\uparrow$ & sAMOTA $\uparrow$ & MOTA $\uparrow$ & MT $\uparrow$ & ML $\downarrow$ & Cost (MB) $\downarrow$ \\
  \hline
  \hline
  CoBEVT~\cite{xu2022cobevt} (*)               & 37.16 & 57.20 & 84.54 & 84.14 & 57.07 & 15.83 & 0.20 \\
  \hline
  Constant covariance      & 41.51 & 55.87 & 89.37 & \textbf{88.68} & 66.67 & 13.67 & \textbf{0.003} \\
  Local BEV feature only   & 43.48 & \textbf{58.05} & 91.48 & 88.46 & 66.91 & 13.43 & 0.0073 \\
  Positional feature only  & 43.45 & 57.85 & 91.44 & 88.38 & 67.63 & 13.67 & 0.0073 \\
  \hline
  DMSTrack (ours)   & \textbf{43.52} & 57.94 & \textbf{91.50} & 88.32 & \textbf{68.35} & \textbf{13.19} & 0.0073 \\
  \hline
\end{tabular}
\end{center}
\vspace{-10pt}
\end{table*}

To figure out which parts of our algorithm contribute to the cooperative tracking performance gain, we conduct ablation experiments on variants of our algorithm, as shown in Table \ref{tab:test_ablation}. Table \ref{tab:test_ablation}'s second row (\textbf{Constant covariance}) shows the result of our multi-sensor Kalman Filter tracking method but with  AB3DMOT~\cite{weng2019ab3dmot}'s default constant observation noise covariance matrix $R$ and initial track state covariance $\Sigma_0$.
We can see that this variant already outperforms V2V4Real~\cite{xu2023v2v4real}'s CoBEVT~\cite{xu2022cobevt} baseline. We also show two more variants that use either only the local BEV feature ${\bf F}_{bev}$ or only the positional feature ${\bf F}_{pos}$ as input of the covariance neural network. We can see that each feature can provide useful information to let the network learn covariance matrices for different detections and achieve better tracking accuracy compared with the CoBEVT~\cite{xu2022cobevt} baseline and our \textbf{Constant covariance} baseline. The best cooperative tracking performance is achieved when both types of features are used.

\subsubsection{Communication Cost}
Our cooperative tracking algorithm's communication cost is closer to V2V4Real~\cite{xu2023v2v4real} \textbf{late fusion}'s cost.
In addition to transmitting $7$ variables per detection box,
our algorithm needs to transmit $10$ more variables of the covariance residual values as shown in Figure \ref{fig:architecture}. Therefore, our communication cost is $\frac{7 + 10}{7} \approx 2.43$ times of the V2V4Real~\cite{xu2023v2v4real}'s \textbf{late fusion}'s cost. So our communication cost is $0.0073$ MB, which is just 0.037 times of the cost of V2V4Real~\cite{xu2023v2v4real}'s CoBEVT~\cite{xu2022cobevt}.

\subsection{Qualitative Results}
\begin{figure}[t!]
    \centering
    \begin{subfigure}[b]{0.48\textwidth}
        \centering 
        \includegraphics[width=\textwidth]{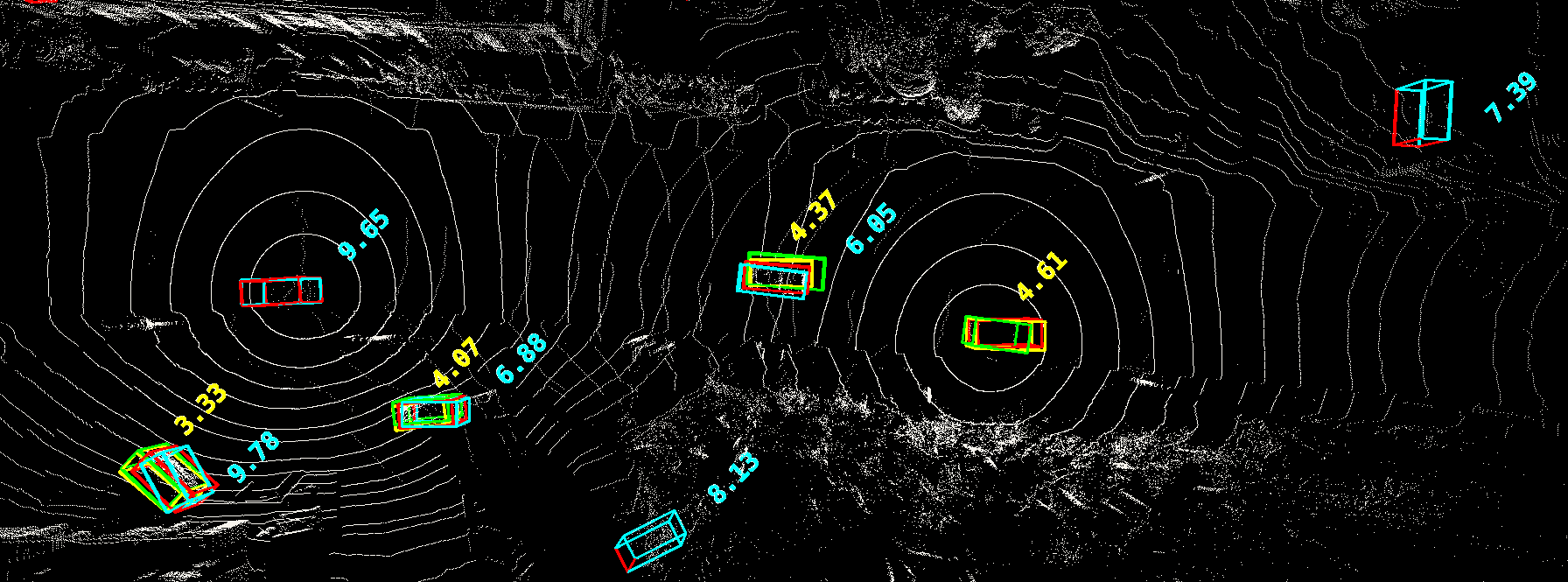}
        \caption[]%
        {{Sequence 3, Frame 2}}    
        \label{fig:seq_0003_frame_0002}
    \end{subfigure}
    \begin{subfigure}[b]{0.48\textwidth}  
        \centering 
        \includegraphics[width=\textwidth]{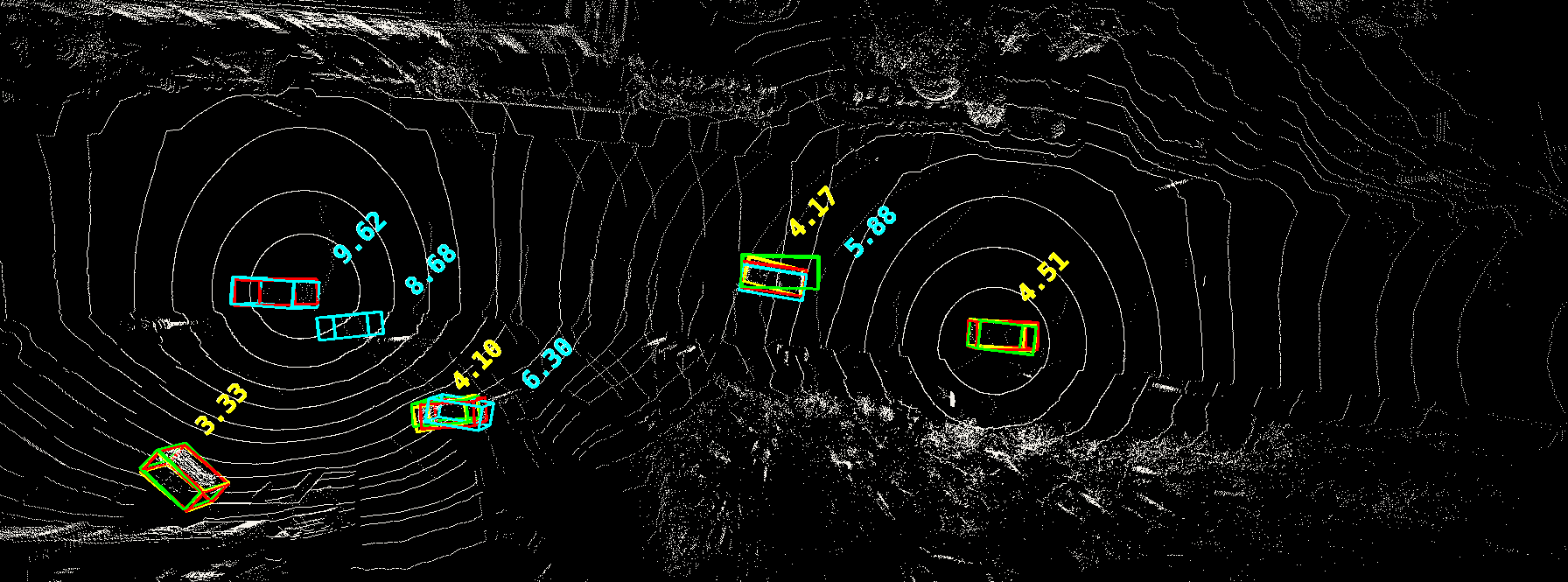}
        \caption[]%
        {{Sequence 3, Frame 3}}    
        \label{fig:seq_0003_frame_0003}
    \end{subfigure}
    \begin{subfigure}[b]{0.48\textwidth}  
        \centering 
        \includegraphics[width=\textwidth]{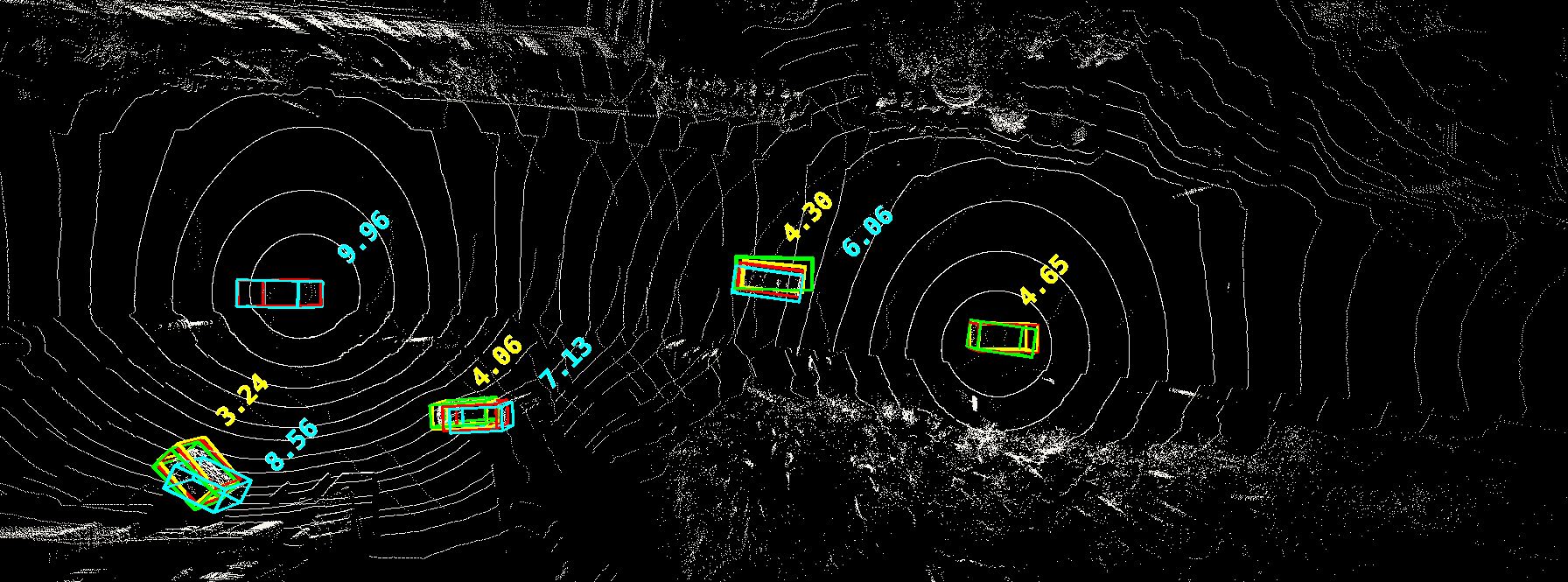}
        \caption[]%
        {{Sequence 3, Frame 4}}    
        \label{fig:seq_0003_frame_0004}
    \end{subfigure}
    \caption[]
    {Qualitative tracking results of our algorithm. 
    \vspace{-5pt}} 
    \label{fig:seq_0003}
\end{figure}

Figure \ref{fig:seq_0003} shows our qualitative results of a testing sequence's $3$ sample frames. \textcolor{green}{Green} and \textcolor{red}{Red} 3D bounding boxes are the ground-truth and the final tracking result. \textcolor{Goldenrod}{Yellow} and \textcolor{cyan}{Cyan} boxes are the individual detections from CAV$1$ and CAV$2$ respectively. CAV$1$ and CAV$2$ are located at the left and right centers of the two circular point cloud patterns respectively. The numbers represent the average of diagonal elements of our model's estimated observation noise covariance matrix ${\bf R}$ of each detection.

For the object in the bottom left corner of Figure \ref{fig:seq_0003_frame_0002} and \ref{fig:seq_0003_frame_0004}, we can see that CAV$1$'s detection (\textcolor{Goldenrod}{Yellow}) is closer to the ground-truth (\textcolor{green}{Green}), compared with CAV$2$'s detection (\textcolor{cyan}{Cyan}). The CAV$2$'s lower detection accuracy on that object is potentially due to partial occlusion and a longer distance from the object to CAV$2$'s sensor.
Our covariance neural network is able to learn a larger observation noise covariance for CAV$2$ on that detection than CAV$1$ given the positional features and the local BEV features. And our differentiable multi-sensor Kalman Filter-based tracking algorithm is able to generate the final tracking result box (\textcolor{red}{Red}) closer to CAV$1$'s detection (\textcolor{Goldenrod}{Yellow}) and the ground-truth (\textcolor{green}{Green}).
Our full-length tracking result videos are available at this \href{https://eddyhkchiu.github.io/dmstrack.github.io/}{URL}.

\section{Conclusion}
We propose a novel probabilistic 3D multi-object cooperative tracking algorithm via a differentiable multi-sensor Kalman Filter. For each detection from each CAV, our model learns to use local BEV features and positional features to estimate observation noise covariance. The estimated uncertainty can be better used to exploit the theoretical property of Kalman Filter-based tracking methods. Our cooperative tracking algorithm achieves better tracking accuracy with lower communication cost in comparison to the state-of-the-art method in V2V4Real~\cite{xu2023v2v4real}. For future work, we would like to explore more advanced tracking algorithms and distributed training approaches: training separate covariance nets independently and chaining them together at tracking inference.


\clearpage
\balance
{\small
\bibliographystyle{IEEEtran}
\bibliography{egbib}
}



\end{document}